\documentclass[lettersize,journal]{IEEEtran}
\usepackage{amsmath,amsfonts}
\usepackage{algorithmic}
\usepackage{algorithm}
\usepackage{array}
\usepackage[caption=false,font=normalsize,labelfont=sf,textfont=sf]{subfig}
\usepackage{textcomp}
\usepackage{stfloats}
\usepackage{url}
\usepackage{verbatim}
\usepackage{graphicx}
\usepackage{cite}
\usepackage[colorlinks=true, citecolor=blue, linkcolor=blue, urlcolor=blue]{hyperref}

\begin{document}

\title{{CondUQ}: Calibrated Conditional Uncertainty Quantification Using Distribution Matching}

\author{Shizhuang Wang\textsuperscript{1,2},~
        Tianhao Wang\textsuperscript{1},~
        Xingqun Zhan\textsuperscript{1,*}\\[6pt]
{
\textsuperscript{1} Shanghai Jiao Tong University, Shanghai, China\\
\textsuperscript{2} Nanyang Technological University, Singapore\\
\textsuperscript{*} Corresponding Author.
E-mail: xqzhan@sjtu.edu.cn}
}


\maketitle

\begin{abstract}

Many applications, especially reliability- and safety-critical ones, require accurate uncertainty quantification for the outputs of model- or learning-based measurement, estimation, and prediction systems. The errors of such systems often vary with factors such as input quality and operating conditions, and thus cannot be accurately described by a single Gaussian model. Although Gaussian mixture models and calibrated quantiles can approximate such error distributions, their unconditional forms do not capture how the error behavior depends on these factors. Hence, we propose \textsc{condUQ}, an approach  that conditions error models on observable factors for enabling accurate and reliable uncertainty quantification. \textsc{condUQ} is built on top of an existing system without modifying it and applies even when the system is a black box. First, \textsc{condUQ} maps the observable factors to a latent quality index (QI) such that errors sharing the same QI follow a common, approximately Gaussian conditional distribution. The QI predictor is trained end-to-end with a distribution-matching 
objective, without requiring quality labels. Second, for each QI, a Gaussian error model is then fitted and calibrated on a large held-out dataset. \textsc{condUQ} is demonstrated on three examples: (a) GPS multipath errors conditioned on satellite elevation and signal strength, (b) relative attitude errors of a transformer-based visual relative pose regression model conditioned on its internal features of the image pairs, and (c) prediction errors of a world model conditioned on the input observations and actions. In all three examples, \textsc{condUQ} substantially reduces the discrepancy between the modeled and empirical error distributions compared with a single global model. Moreover, the learned QIs reveal how the underlying factors affect the error characteristics in a manner consistent with physical principles.

\end{abstract}

\begin{IEEEkeywords}
Uncertainty quantification, conditional error modeling, distribution matching, unsupervised learning
\end{IEEEkeywords}

\section{Introduction}

Uncertainty quantification (UQ) is highly essential for many applications, especially safety- or reliability-critical ones, such as autonomous driving~\cite{wang2025autonomous}, aviation~\cite{wang2021aircraft}, and robotics~\cite{chen2025uncertainty}. These applications require accurate quantification of the uncertainty in the outputs of their measurement, prediction, and estimation systems, because the outputs and their associated  uncertainties may both be used to make critical decisions. For example, an autonomous vehicle uses the output of its perception system to determine whether there is enough clearance between itself and the nearby obstacles \cite{chen2025hyperdimensional}. Measurement noise of the camera or LiDAR data can cause the perception output to differ from the truth. If the output uncertainty is not accurately quantified, then perception error can potentially lead to a safety risk. Thus, uncertainty quantification helps the system determine when an output can be safely used and how much how much confidence that can be placed in it \cite{bickfordsmith2025rethinking}. Also, UQ helps understand the quality of a measurement, prediction, and estimation system and offers guidance for improving the performance.

\begin{figure}[!t]
    \centering
    \includegraphics[width=0.95\columnwidth]{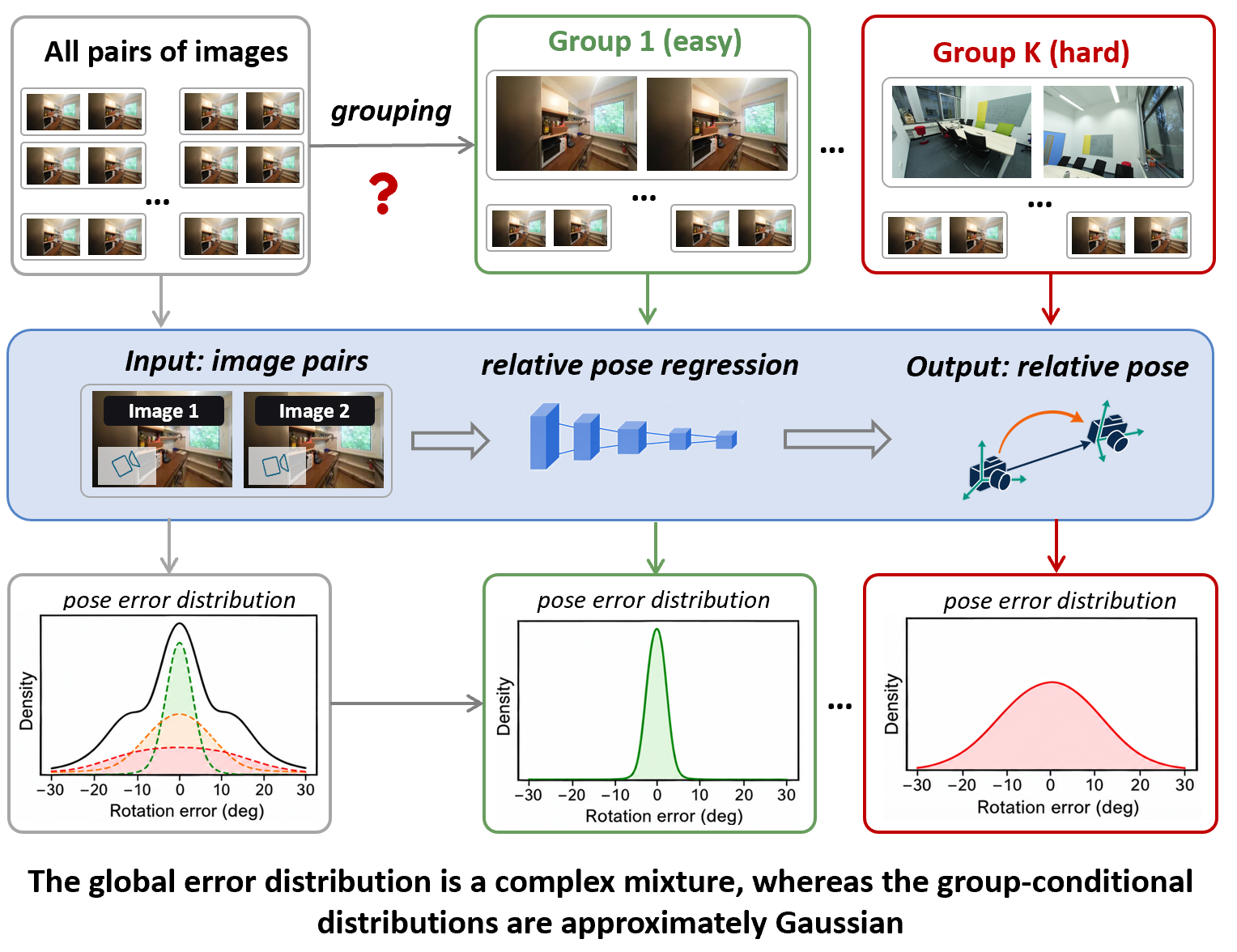}
    \caption{Motivation for conditional uncertainty quantification in relative pose estimation. Image pairs with different feature richness and overlap exhibit different error behavior. Pooling all these errors produces a complex global distribution that is poorly represented by a single Gaussian. Grouping samples by a latent quality index (for example, from easy to hard) yields approximately Gaussian conditional error distributions. {CondUQ} learns to predict this index from observable inputs without quality annotations.}
    \label{fig:motivation}
\end{figure}

\begin{figure*}[!t]
    \centering
    \includegraphics[width=0.9\linewidth]{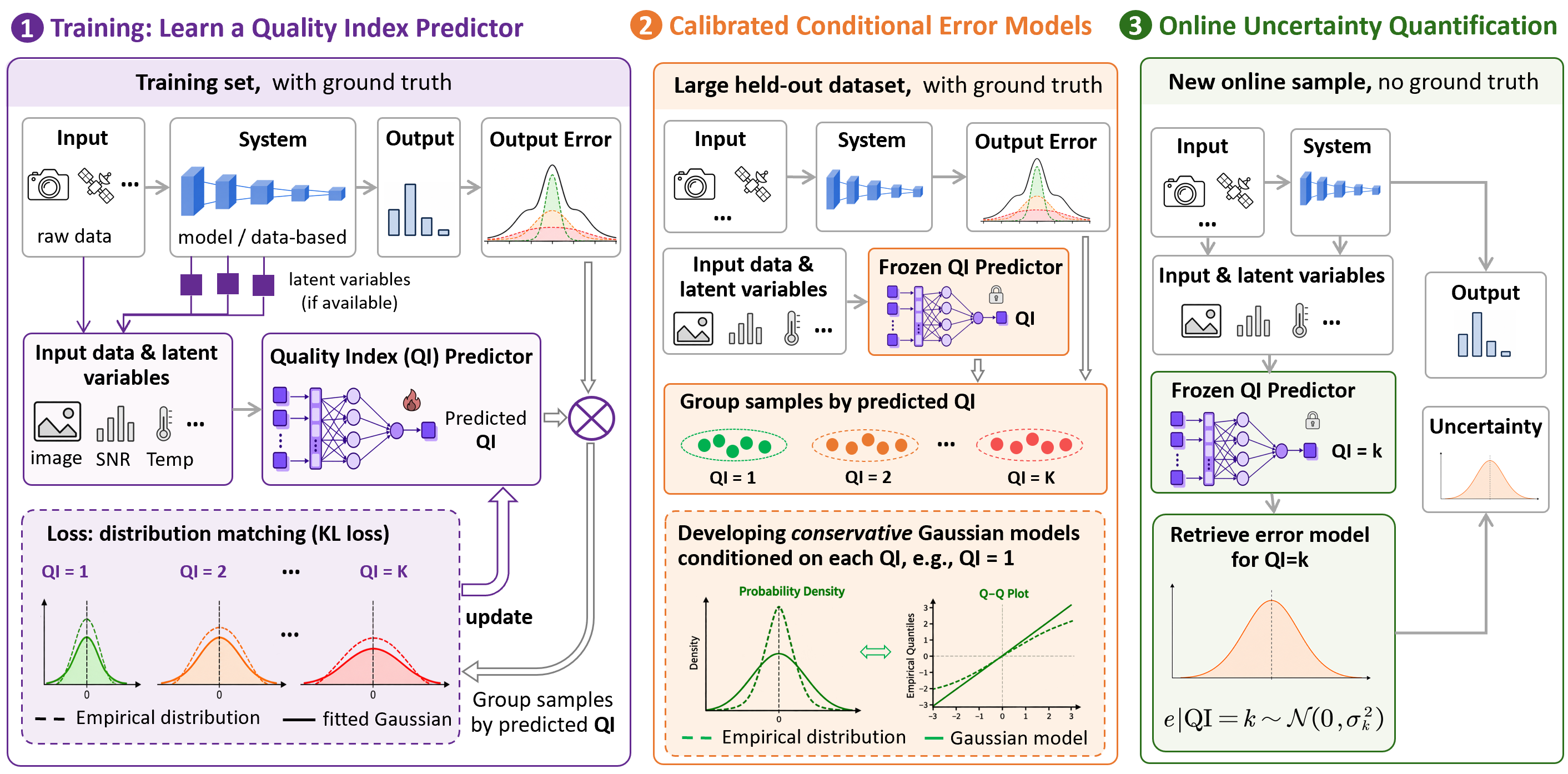}
    \caption{Overview of the proposed \textsc{condUQ} framework. First, a quality index (QI) predictor is learned from input-dependent data (e.g., image pairs) and latent variables in the pipeline of the measurement, estimation, or prediction system. Without QI annotations, a distribution-matching loss encourages the conditional error distribution associated with each predicted QI to be approximately Gaussian. Second, samples in a large held-out calibration dataset are partitioned into groups, with each group comprising samples assigned the same QI by the frozen QI predictor. A conservative Gaussian error model is then fitted separately to each group. Third, at inference time, the predicted QI of a new sample selects the corresponding conditional error model, providing calibrated uncertainty for the system output without requiring ground truth.}
    \label{fig:overview}
\end{figure*}

The output error distribution of a measurement, prediction, or estimation system typically varies with factors such as input quality and operating conditions. Observable factors associated with these conditions can therefore provide useful information about error behavior. In GPS positioning, for example, signal multipath errors vary with signal environment: they are usually small when a receiver is operating in a open-sky condition but can be large in urban areas. The elevation angle and received signal strength of a tracked satellite often serve as observable factors that indicate the behavior of its multipath errors. Also, for relative pose estimation from image pairs, the pose error depends on the richness of visual features in each image and the overlap between the two images. For both examples, since the error behaviors vary with multiple factors, constructing a global error model that describes the error distribution in an average sense is not an ideal solution for accurate uncertainty quantification. \textbf{Fig}. \ref{fig:motivation} illustrates this point using relative pose estimation as an example. As shown in this figure, pooling all error samples across different image-pair conditions produces a complex distribution that is poorly represented with a single Gaussian model. However, grouping samples with similar error behavior yields simpler, approximately Gaussian distributions within each group. This observation motivates our framework, \textsc{condUQ}, which aims to learn a latent quality index (QI) from observable inputs, without requiring quality annotations. The QI predictor is learned using a distribution matching objective, such that those samples sharing the same QI follow a similar, approximately Gaussian conditional distribution. Second, for each quality index, \textsc{condUQ} uses a large held-out dataset to fit and calibrate a Gaussian error model. Finally, during online uncertainty quantification, the QI is predicted from observable inputs and used to select the corresponding error model. This framework is illustrated in \textbf{Fig}. \ref{fig:overview}. 

The main contributions of this work are as follows:
\begin{itemize}[\setlength{\IEEElabelindent}{0pt}]
\item We introduce a distribution-matching objective that learns a latent quality index from the observable inputs and internal variables of a system without requiring quality labels. This objective encourages the samples sharing the same quality index (QI) to have the output errors that follow a common, approximately Gaussian conditional distribution.
\item We develop \textsc{condUQ}, a UQ framework that combines QI prediction with fitting and calibrating a separation Gaussian error model for each QI. At inference time, the QI and its corresponding error model are obtained from the inputs and internal variables of the system, without the need to change the system or get the ground truth.
\item We evaluate \textsc{condUQ} on three examples: GPS multipath errors, image-based relative pose errors, and next-step state prediction errors of a world model. In all three applications, our approach yields better agreement between the modeled and empirical error distributions than a single global error model. Further analysis shows that the learned relationships between the conditioning factors and the error behavior are consistent with known physical principles.
\end{itemize}

\section{Related Work}

Existing approaches for uncertainty quantification (UQ) can be grouped into four categories according to their outputs and how they account for the dependence of the error on the input: sample-level uncertainty scores, calibrated uncertainty  output, conformal prediction, and conditional error modeling. In this article, we focus on characterizing the output error of a fixed system under the operating conditions covered by the collected dataset. Out-of-distribution (OOD) cases are beyond the scope of this work, although \textsc{condUQ} offers guidance to identify such cases. OOD will be addressed in future work.

\subsection{Sample-Level Uncertainty Scores}

A large body of work assigns a scalar uncertainty score to each output of a system. One line of work predicts the score using information learned during training. For example, \cite{{li2026reliable}} uses image features learned from data to predict uncertainty of each classification result; \cite{apostolopoulou2024rate} uses the distance from a new sample to a learned compressed representation of the training data space to measure the uncertainty of the prediction output. The second line of work estimates the uncertainty of an output from the sensitivity of the prediction result to perturbations of the network parameters \cite{wilson2025empirical}, the intermediate features \cite{xu2026cupid}, or the input images \cite{hornauer2025revisiting}. The third line measures the difference among the predictive distributions from multiple models \cite{berry2025pairwise}. These scores are useful to give an insight of the uncertainty or detect OOD inputs. However, a scalar score offers limited information about the full error distribution, and it carries no calibration guarantee, i.e., a given score does not correspond to a known error distribution or magnitude range.

\subsection{Calibrated Uncertainty Quantification}

Calibration UQ approaches use held-out data to correct the predictive distributions or quantiles of an existing UQ model. For example, \cite{kuleshov2018calibrated} corrects the predicted quantiles so that they match the empirical values computed from the held-out data; \cite{song2019distribution} takes the predicted mean and variance of each sample as inputs to a calibrator, which estimates adjustment parameters to correct the original distributions. These methods ensure that the distributions or quantiles are calibrated on average over the entire input space, but samples under different conditions (as shown in \textbf{Fig}. \ref{fig:motivation}) can still receive the same correction, so the calibrated distribution or quantile may differ from the true one under specific conditions. To address this limitation, \cite{dey2025ladar} aims to learn a local probability--probability map that checks where an initial conditional distribution estimate is miscalibrated and reshapes it toward instance-wise calibration. However, such an approach requires the system to already provide a predictive conditional distribution. 

\subsection{Conformal Prediction}

Conformal prediction is a framework for UQ that produces prediction intervals or regions under a specified coverage level (e.g., 99\%) for any predictive model \cite{angelopoulos2023conformal}. In \cite{papadopoulos2024guaranteed},  conformal prediction (CP) is combined with Gaussian process regression to produce guaranteed prediction intervals. In addition, \cite{xu2024ellipsoidal} constructs ellipsoidal prediction regions for multivariate time series. In \cite{kondratyev2026neural}, neural optimal transport is utilized to learn the conditional quantiles of multivariate errors; the quantiles are then used to construct the prediction regions. These methods guarantee the coverage level in an average sense but may hide undercoverage under specific conditions. \cite{jung2023batch} strengthens the guarantee to hold conditionally on each predefined group of samples, but the groups must be defined in advance. Instead, \cite{cabezas2024regression} learns the group partition from data: it trains a regression tree on conformity scores, then uses its leaves to partition the feature space, and derives a separate conformal threshold for each group. Because the tree separate samples by the values of their scores, the resulting groups are not designed to make the group-conditioned distribution approximately Gaussian. In addition, this method operates on tabular features and outputs prediction intervals rather than full distributions. Also, more generally, CP methods describe the range of possible errors at a given coverage level without specifying the error distribution within that range.

\subsection{Conditional Error Modeling}

In addition to conditional conformation prediction in \cite{cabezas2024regression}, there are several approaches about conditional error modeling. These methods capture the fact that error behaviors can vary with input quality and operating conditions. Traditional error modeling often pools error samples from all input conditions and fits a single global distribution, such as a Gaussian \cite{blanch2019gaussian} or a Gaussian mixture \cite{wang2026gmm}. Even when such a model fits the global distribution well, it cannot reflect how the error behavior changes across conditions. A common remedy is to divide the manually-selected features into intervals based on experience and build a separate model for each interval. For example, in GPS positioning for aviation, airborne GPS multipath errors are modeled separately for different satellite elevation intervals \cite{circiu2020multipath}. However, such partitions rely heavily on prior experience to set the boundaries, may overlook other features, and become difficult to design as the number of features grows. Learning-based methods remove the need for manual partitions through training a network to predict the error distribution or quantile of each sample by using its input, output, and other observable variables. Mixture density networks~\cite{bishop1994mixture} predict a conditional distribution for each sample using a Gaussian mixture, so it is a sample-level conditional UQ method without calibration. In addition, \cite{kendall2017uncertainties} builds upon Bayesian neural network and trains a network to predict a variance for each sample, which defines the Gaussian distribution of the output error. To improve UQ performance,  \cite{mao2026sparse} predicts both the global noise variance and the input-dependent variance with a Gaussian process model. Besides, \cite{kostic2024neural} learns the relationship between input and output features and uses it to adjust the conditional error distribution under the current input. All these methods learn a per-sample conditional error distribution without incorporating an explicit calibration step. A recent work \cite{sosamarco2025contextualised} learns a normalizing flow on top of a trained model, which produces error distributions conditioned on the model output and internal features. It gives good empirical calibration performance, but its calibration is not guaranteed since no calibration set is used. For the methods above, even if a calibration step were added, the calibration can only be achieved over the entire dataset because each predicted distribution is only associated with one error sample. Thus, the predicted distributions may be nicely calibrated on average but can be miscalibrated under specific conditions.

\subsection{Novelty of This Approach}

The proposed method \textsc{CondUQ} offers accurate uncertainty quantification using a calibrated conditional distribution. This approach differs from the above methods in three main aspects. First, it learns how to group samples without labels or manual design. It instead directly learns the quality index (QI) through a distribution-matching objective, so that the errors within each QI follow a common, approximately Gaussian distribution. QI is therefore a latent representation of the operating condition that directly relates to the error behavior. Second, it fits and calibrates the error models under each QI  rather than only on average. Because there are many samples associated with each QI, calibration is reliable, whereas this is not the case for the methods that focus on per-sample error distributions. Third, it provides a full conditional distribution for 
the system's output error using Gaussian representations, which is easy to use and interpret. Also, since the QI predictor can be any differential network and the system can be treated as a full or partial black box, \textsc{condUQ} applies to model- and learning-based systems with inputs of arbitrary modality.

\section{Methodology}

\label{sec:method}

We propose \textsc{CondUQ}, a calibrated conditional uncertainty quantification framework that can be built on top of any model /learning-based measurement, estimation, or prediction system without modifying it. As shown in \textbf{Fig}.
\ref{fig:motivation}, a complex global error distribution may result from mixing heterogeneous operating conditions, whereas errors under similar conditions often follow simpler distributions. Therefore, \textsc{condUQ} introduces a discrete quality index (QI) to group samples that have similar error behavior, enabling more accurate UQ than a single global error model. As shown in \textbf{Fig}. \ref{fig:overview}, \textsc{condUQ} consists of three stages: (1) learning a QI predictor, (2) fitting and calibrating a conditional error model for each QI, and (3) performing an online uncertainty quantification.

\subsection{Problem Formulation and Framework Overview}
\label{sec:setup}

\textbf{System.} Let $\mathbf{u}\in\mathbb{U}$ denote a random input, and let
$\mathbf{y}^{*}\in\mathbb{R}^n$ be the corresponding ground-truth output. The input
$\mathbf{u}$ may be multimodal, for example, images, signals, or text. For an observed input sample $\boldsymbol{u}_i$, a fixed
measurement, estimation, or prediction system $f$ produces the output
\begin{equation}
    \hat{\boldsymbol{y}}_i=f(\boldsymbol{u}_i)\in\mathbb{R}^n.
    \label{eq:system}
\end{equation}
Equivalently, the system induces the random output
$\hat{\mathbf{y}}=f(\mathbf{u})$. We impose no structural or differentiability
assumptions on $f$: it may be model-based or learned and may be fully known,
partially known, or accessible only as a black box. The system may also expose observable internal variables. Let $\mathbf{h}$ denote these random 
variables and $\boldsymbol{h}_i$ their realization for sample $i$. Noted that $\boldsymbol{h}_i$ is available without the ground truth $\boldsymbol{y}_i^*$. For example, in relative pose regression, $\boldsymbol{u}_i$ is an
image pair, $\hat{\boldsymbol{y}}_i$ is the estimated relative pose, and $\boldsymbol{h}_i$ may
contain intermediate features from the regression network.

\textbf{Objective.} Our goal is to enable accurate and reliable online uncertainty quantification for the system output. The random output error is given as 
\begin{equation}
    \mathbf{e}=\varepsilon(\hat{\mathbf{y}},\mathbf{y}^{*})
    \label{eq:error}
\end{equation}
where $\varepsilon$ is an application-dependent error function, e.g., we have $\mathbf{e}=\hat{\mathbf{y}}-\mathbf{y}^{*}$ for an additive error. We aim to characterize the behavior of the output error during online operation where the ground-truth output is unavailable. The output error $\mathbf{e}$ may follow a complex non-Gaussian distribution because it pools errors under different operating conditions (\textbf{Fig}. \ref{fig:motivation}), i.e.,
\begin{equation}
    p(\mathbf{e})=\sum_{c=1}^{C} p(\mathbf{e} \mid \mathcal{C}=c)P(\mathcal{C}=c)
    \label{eq:global}
\end{equation}
where $\mathcal{C}=c$ represents the $c$-th operating condition. The conditional distribution $p(\mathbf{e} \mid \mathcal{C}=c)$ is often simpler than the global distribution $p(\mathbf{e})$; for example, it can be approximately Gaussian. We therefore introduce a discrete quality index $\mathrm{q}$ to be a latent representation of the operation condition, i.e.,
\begin{equation}
    \mathrm{q}=\phi(\mathbf{u},\mathbf{h}) \in \{1,\ldots,K\}, \quad q_i=\phi(\boldsymbol{u}_i,\boldsymbol{h}_i)
\end{equation}
where $\phi(\mathbf{u},\mathbf{h})$ denotes the quality index (QI) predictor, $K$ is a predefined hyperparameter, and $q_i$ is the QI assignment of sample $i$. We seek to learn a QI predictor $\phi$ such that the error distribution conditioned on each QI is approximately Gaussian:
\begin{equation}
    p(\mathbf{e}\mid\mathrm{q}=k)
    \approx
    \mathcal{N}(\mathbf{e};\boldsymbol{\mu}_k,\boldsymbol{\Sigma}_k),
    \qquad
    k=1,\ldots,K,
    \label{eq:conditional_gaussian}
\end{equation}
where $\boldsymbol{\mu}_k$ and $\boldsymbol{\Sigma}_k$ denote
the mean and covariance matrix associated with QI=$k$, respectively. This formulation replaces a complex global error distribution with a collection of simpler QI-conditional distributions, enabling a more accurate UQ, as shown in \textbf{Fig}. \ref{fig:motivation}.

\textbf{\textsc{condUQ} Framework.} We assume access to a sufficiently large offline dataset $\mathbb{D}_{\mathrm{off}}$; it is randomly divided into disjoint training and calibration sets. The two sets include ground-truth outputs to compute the output errors. Using the two datasets, the proposed \textsc{condUQ} realizes the above objective in three stages. \textit{First}, using the pairs $(\boldsymbol{u}_i, \boldsymbol{h}_i, \boldsymbol{e}_i)$ in the training set, a QI predictor $\phi$ is learned without QI labels. The distribution-matching loss encourages the samples sharing the same QI to have the output errors that follow a common, approximately Gaussian distribution. \textit{Second}, the learned QI predictor
is then frozen and applied to the calibration set. The samples in this set are
grouped according to their predicted QIs. A calibrated conditional error model is derived separately for each group. 
\textit{Finally}, for a new online
input $\boldsymbol{u}_{\star}$, the fixed system produces
$\hat{\boldsymbol{y}}_{\star}=f(\boldsymbol{u}_{\star})$ and the available internal
variables $\boldsymbol{h}_{\star}$. Using $\boldsymbol{u}_{\star}$ and $\boldsymbol{h}_{\star}$, the output of the frozen QI predictor is QI=$\hat{q}_{\star}$, which selects the corresponding calibrated
conditional error model. The selected model can characterize the unknown error of
$\hat{\boldsymbol{y}}_{\star}$, thereby providing calibrated uncertainty without
requiring the ground-truth output $\boldsymbol{y}_{\star}^{*}$.


\subsection{Training a Quality Index Predictor}
\label{sec:qi_predictor}

We aim to learn a QI predictor using the observable pairs $(\boldsymbol{u}_i,\boldsymbol{h}_i)$ and the output errors $\boldsymbol{e}_i$ computed from the ground truth in the training set. Instead of using QI annotations, a distribution-matching loss is utilized to encourage the error distribution associated
with each predicted QI to approach an approximately Gaussian distribution.

\textbf{QI Predictor.}
The QI predictor can use any differentiable network architecture that is appropriate for the available inputs. For example, an MLP can be used for vector-valued inputs, while convolutional networks or Transformers can be utilized for images, signals, text, or multi-modal inputs. In general, a network $g_{\boldsymbol{\theta}}$ produces
$K$ logits:
\begin{equation}
    \boldsymbol{z}_i=g_{\boldsymbol{\theta}}(\boldsymbol{u}_i,\boldsymbol{h}_i)\in\mathbb{R}^K,
    \label{eq:qi_logits}
\end{equation}
where $\boldsymbol{\theta}$ denotes the learnable parameters. If internal variables are
unavailable, then $\boldsymbol{h}_i$ is omitted. During training, the logits are
converted into soft assignment probabilities to make the QI assignment differentiable:
\begin{equation}
    \pi_{ik}
    =
    \frac{\exp(z_{ik}/\tau)}
    {\sum_{j=1}^{K}\exp(z_{ij}/\tau)},
    \qquad k=1,\ldots,K,
    \label{eq:soft_qi}
\end{equation}
where $\pi_{ik}$ is the probability of assigning sample $i$ to QI $=k$, and
$\tau>0$ is a temperature parameter. We decrease $\tau$ during training:
a high initial temperature $\tau_0$ facilitates optimization through smooth
assignments, whereas a low final temperature makes the assignments approach
one-hot vectors. And once the network has been trained, its parameters are frozen and the soft assignment is replaced by the deterministic QI predictor as follows \begin{equation} \hat{q}_i = \phi_{\boldsymbol{\theta}}(\boldsymbol{u}_i,\boldsymbol{h}_i) = \operatorname*{arg\,max}_{k\in\{1,\ldots,K\}}z_{ik}. \label{eq:hard_qi} \end{equation} 
The frozen predictor $\phi_{\boldsymbol{\theta}}$ is then applied to the calibration and online samples.

\textbf{Distribution-Matching Loss.}
To train the QI predictor, we define a distribution-matching loss to
measure the Gaussianity of the error distribution associated with each QI. Note that the error $\mathbf{e}\in\mathbb{R}^n$ may be scalar ($n=1$) or multivariate ($n>1$). For a
scalar error, Gaussianity is measured directly using the KL divergence
between its empirical distribution and a fitted univariate Gaussian. For a
multivariate error, however, it is computationally expensive to directly compute the KL divergence between multidimensional
distributions. Therefore,  we measure multivariate Gaussianity through multiple
random one-dimensional projections. That is, at each training iteration, we randomly select $m$ unit-norm projection directions that are sampled uniformly on the hypersphere, denoted by $\{\mathbf{a}_r\}_{r=1}^{m}$. For $n=1$, we simply set $m=1$ and $\mathbf{a}_1=1$. For each sample $i$ and each direction $\mathbf{a}_r$, we project the error into a scalar:
\begin{equation}
    s_{ir}
    =
    \mathbf{a}_r^{\top}\boldsymbol{e}_i.
    \label{eq:projected_error}
\end{equation}
For QI $=k$, we define its effective sample size using the soft assignments $\pi_{ik}$ as
\begin{equation}
    d_k
    =
    \sum_{i=1}^{N_{\mathrm{tr}}}\pi_{ik}
    \label{eq:effective_group_size}
\end{equation}
where $N_{\mathrm{tr}}$ is the number of training samples. Then, for each direction $\mathbf{a}_r$, we fit a
univariate Gaussian using the weighted mean and variance
\begin{equation}
    \widehat{\mu}_{kr}=
    \frac{1}{d_k}
    \sum_{i=1}^{N_{\mathrm{tr}}}
    \pi_{ik}s_{ir}, \quad
    \widehat{\sigma}_{kr}^{2}=
    \frac{1}{d_k}
    \sum_{i=1}^{N_{\mathrm{tr}}}
    \pi_{ik}
    (s_{ir}-\widehat{\mu}_{kr})^2.
    \label{eq:projected_variance}
\end{equation}
Let $\{\mathcal{B}_{rb}\}_{b=1}^{B}$ denote a fixed, 
$\boldsymbol{\theta}$-independent partition of the projected error space along the direction
$\mathbf{a}_r$. The weighted empirical bin probabilities are
\begin{equation}
    \widehat{p}_{krb}
    =
    \frac{1}{d_k}
    \sum_{i=1}^{N_{\mathrm{tr}}}
    \pi_{ik}\,
    \mathbb{I}\{s_{ir}\in\mathcal{B}_{rb}\},
    \qquad
    b=1,\ldots,B.
    \label{eq:projected_histogram}
\end{equation}
Next, we measure the Gaussianity of the error distribution for QI $=k$ along the direction $\mathbf{a}_r$ using the KL divergence as
\begin{equation}
    D_{kr}
    =
    \sum_{b=1}^{B}
    \widehat{p}_{krb}
    \log
    \frac{\widehat{p}_{krb}}
         {\widehat{g}_{krb}}
    \label{eq:projected_kl}
\end{equation}
where $\widehat{g}_{krb}$ denotes the bin probability assigned to $\mathcal{B}_{rb}$ by $\mathcal{N}(\widehat{\mu}_{kr},\widehat{\sigma}_{kr}^2)$. A smaller $D_{kr}$ indicates that the projected errors along $\mathbf{a}_r$
assigned to QI $=k$ are better approximated by a Gaussian. The distribution-matching loss sums the projected KL divergences across
all QIs and projection directions:
\begin{equation}
    \mathcal{L}_{\mathrm{G}}(\boldsymbol{\theta})
    =
    \frac{1}{mK}
    \sum_{r=1}^{m}
    \sum_{k=1}^{K}
    (1+D_{kr})^2.
    \label{eq:gaussianity_loss}
\end{equation}
This construction is motivated by Cramér--Wold, i.e.,
a random vector is a multivariate Gaussian if all of its one-dimensional
projections are all Gaussian. Because the sampled directions, projected errors,
and bin assignments are independent of $\boldsymbol{\theta}$, the loss remains
differentiable with respect to the soft assignments $\pi_{ik}$ and hence the
QI predictor parameters.

\textbf{Balance Regularization.}
When trained by optimizing only $\mathcal{L}_{\mathrm{G}}$, the QI predictor may assign most samples to only a few QIs, leaving the remaining QIs sparsely populated or empty. We therefore define the marginal QI proportions as
\begin{equation}
    \bar{\pi}_k
    =
    \frac{d_k}{N_{\mathrm{tr}}},
    \qquad k=1,\ldots,K,
\end{equation}
and we encourage balanced assignments by maximizing their normalized entropy:
\begin{equation}
    \mathcal{H}_{\mathrm{QI}}(\boldsymbol{\theta})
    =
    -\frac{1}{\log K}
    \sum_{k=1}^{K}
    \bar{\pi}_k\log\bar{\pi}_k.
    \label{eq:balance_entropy}
\end{equation}
The complete training objective is therefore given as
\begin{equation}
    \mathcal{L}_{\mathrm{QI}}(\boldsymbol{\theta})
    =
    \mathcal{L}_{\mathrm{G}}(\boldsymbol{\theta})
    -
    \eta\cdot\mathcal{H}_{\mathrm{QI}}(\boldsymbol{\theta}),
    \label{eq:qi_training_loss}
\end{equation}
where $\eta\geq0$ is a hyperparameter that controls the balance regularization. We gradually decrease $\eta$ from $\eta_0$ during training to prevent early group
collapse while allowing the distribution-matching loss to dominate at later
stages.

\subsection{Calibrating QI-Conditional Error Models}
\label{sec:conditional_calibration}

After training, the QI predictor is frozen and then applied to the calibration set to develop conditional error models for each QI. For each QI $=k$, we collect the corresponding errors as follows
\begin{equation}
    \mathcal{E}_k
    =
    \left\{
        \boldsymbol{e}_i:
        \phi_{\boldsymbol{\theta}}(\boldsymbol{u}_i,\boldsymbol{h}_i)=k,\,
        (\boldsymbol{u}_i,\boldsymbol{h}_i,\boldsymbol{e}_i)\in\mathbb{D}_{\mathrm{cal}}
    \right\}
    \label{eq:calibration_errors}
\end{equation}
where $\mathbb{D}_{\mathrm{cal}}$ denotes the calibration set. The calibration set is independent of the training set and is assumed to contain a sufficiently large number of samples.

\textbf{Modeling Objective.}
We aim at constructing a conditional error model for each QI, which can accurately
represent the empirical error distribution while remaining simple to use and interpret.
 For each QI $=k$, let us consider an $M$-component Gaussian mixture model
\begin{align}
    p_{k,M}(\mathbf{e};\Theta_{k,M})
    &=
    \sum_{\ell=1}^{M}
    \omega_{k\ell}
    \mathcal{N}
    \left(
        \mathbf{e};
        \boldsymbol{\mu}_{k\ell},
        \boldsymbol{\Sigma}_{k\ell}
    \right), \nonumber  \\
    \quad \omega_{k\ell}&\geq0, \quad \sum_{\ell=1}^{M}
    \omega_{k\ell} = 1
    \label{eq:conditional_gmm}
\end{align}
where $\omega_{k\ell}$ denotes the mixture weight and $\Theta_{k,M}$ collects all the mixture
weights, means, and covariance matrices. The case $M=1$ corresponds to a
single Gaussian distribution. We use a single Gaussian distribution in most cases and only consider a Gaussian mixture when necessary. We start from $M=1$, i.e., using a single Gaussian model as the conditional model. We then seek the best model parameters to minimize the following distribution-matching loss
\begin{equation}
    \mathcal{L}_{k,M}(\Theta_{k,M})
    =
    \frac{1}{m}
    \sum_{r=1}^{m}
    \left(
        1+D_{kr,M}(\Theta_{k,M})
    \right)^2,
    \label{eq:calibration_matching_loss}
\end{equation}
where $D_{kr,M}(\Theta_{k,M})$ denote the KL divergence between the model and the empirical distributions for QI $=k$ along the unit-norm direction $\mathbf{a}_r$. Note that the $m$ projection directions are sampled uniformly on the hypersphere. We set $m=1$ and $\mathbf{a}_1=1$ for scalar errors. Let $\mathcal{L}_{k,M,\mathrm{opt}}$ be the minimum loss and $\Theta^{}_{k,M,\mathrm{opt}}$ be the corresponding best model. Then this model is accepted as the conditional error model for QI $=k$ if the following criterion is met
\begin{equation}
\mathcal{L}_{k,M,\mathrm{opt}} = \mathcal{L}_{k,M}(\Theta_{k,M,\mathrm{opt}}) < T,
    \label{eq:loss_criterion}
\end{equation}
where $T$ is a predefined threshold for KL divergence. If this is not met, we increase the number of Gaussian components, $M$, by one and repeat the
optimization until \eqref{eq:loss_criterion} is satisfied. In short, we seek the error model that has the fewest Gaussian components while meeting the fitting accuracy (\ref{eq:loss_criterion}). Benefiting from the training process of the QI predictor, a single Gaussian model is sufficient in most cases. Finally, let us denote the resulting conditional error model for QI $=k$ by
\begin{equation}
    p(\mathbf{e}\mid\mathrm{q}=k)
    \approx
    p
    \left(
        \mathbf{e};
         {\Theta}_{k,M,\mathrm{cal}}
    \right).
    \label{eq:calibrated_conditional_model}
\end{equation}
Specifically, if the resulting error model has a single Gaussian component (i.e., $M=1$), we have
\begin{equation}
    p(\mathbf{e}\mid\mathrm{q}=k)
    \approx \mathcal{N}(\mathbf{e};\boldsymbol{\mu}_{k,\mathrm{cal}},\boldsymbol{\Sigma}_{k,\mathrm{cal}}).
    \label{eq:calibrated_conditional_model_gaussian}
\end{equation}
This procedure favors a single Gaussian distribution whenever it adequately represents the
calibration errors and introduces additional mixture components only when
required.

\textbf{Parameter Optimization.} For a fixed $M$, the parameters $\Theta_{k,M}$ of the Gaussian mixture model can be optimized using gradient descent. The mixture weights
are parameterized with a softmax, and each covariance matrix is
parameterized using a regularized Cholesky factor. Because the objective is
nonconvex for $M>1$, we should use multiple initializations and retain the solution
with the smallest loss. For multivariate errors, projection directions are
resampled during parameter optimization process.

\subsection{Online Uncertainty Quantification}
\label{sec:online_uq}

During online operations, the uncertainty of the system's output can be quantified using the frozen QI predictor and the calibrated conditional error models. Given a new online input $\boldsymbol{u}_{\star}$, the fixed system produces
$\hat{\boldsymbol{y}}_{\star}=f(\boldsymbol{u}_{\star})$ and, when available, the internal variables
$\boldsymbol{h}_{\star}$. The frozen QI predictor assigns the sample to
$\hat{q}_{\star}=\phi_{\boldsymbol{\theta}}(\boldsymbol{u}_{\star},\boldsymbol{h}_{\star})$, which selects the
corresponding error model:
\begin{equation}
    p(\mathbf{e}_{\star}\mid\boldsymbol{u}_{\star},\boldsymbol{h}_{\star})
    =
    {p}(\mathbf{e}_{\star}\mid q=\hat{q}_{\star})
    \label{eq:online_error_distribution}
\end{equation}
where the QI-conditional error model is given in (\ref{eq:calibrated_conditional_model}) and (\ref{eq:calibrated_conditional_model_gaussian}). The selected model characterizes the unknown error of
$\hat{\boldsymbol{y}}_{\star}$ and can be used to report moments, quantiles, or uncertainty
regions without requiring the ground-truth output
$\boldsymbol{y}_{\star}^{*}$.

\section{Illustrative Examples and Discussions}

We evaluate \textsc{condUQ} on three illustrative examples: GPS multipath errors conditioned on satellite elevation and signal strength, learning-based relative pose regression (reloc3R \cite{Dong_2025_CVPR}) conditioned on the input image pairs, and prediction errors of a world model (LeWM \cite{maes2026leworldmodel}) conditioned on the observations and actions. The results are presented in the following parts, and discussions are also provided.

\subsection{GPS Multipath Errors}
\label{secIII-A}

This example considers a measurement system, i.e., a GPS receiver, and models the distribution of the multipath errors in GPS range measurements. The QI predictor takes the elevation and the carrier-to-noise density ratio (C/N0) of each tracked satellite as inputs, both of which are scalars. The two inputs are available from the receiver and are known to be related to multipath errors. This example represents a case in which the input space for the QI predictor is only 2-D, and it allows the learned QIs to be directly interpreted with the two physically meaningful features. The multipath error itself is 1-D.

The dataset consists of GPS observations collected at IGS stations from January 1 to 30, 2024, with a sampling interval of 10~s. Multipath errors are computed using an open-source software \cite{aarnesGnssMultipath}. Only samples with elevation angles of at least 5 degrees are used to construct the dataset, which includes 10$^6$ samples in total. The dataset is then randomly divided into a training set (70\%) and a calibration set (30\%). A lightweight MLP is trained as the QI predictor with the number of QI values $K=6$. The resulting QIs are labeled G1 through G6 in ascending order of the variance fitted on the training set. The number of parameters in the MLP is 98,304. The training phase takes only about 2 minutes on a server with a GeForce RTX 5090 GPU (32GB).

\begin{figure}[!t]
    \centering
    \includegraphics[width=0.9\columnwidth]
    {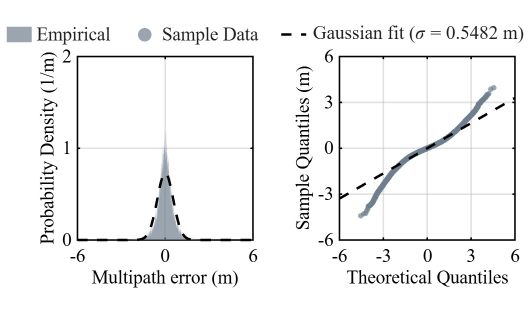}
    \caption{Global distribution of GNSS multipath errors. Left: empirical and fitted Gaussian densities. Right: quantile-quantile (QQ) plot, i.e., the empirical quantiles against standard normal quantiles.}
    \label{fig:GNSS_GLOBAL}
\end{figure}

\begin{figure}[!t]
    \centering
    \includegraphics[width=0.65\columnwidth]
    {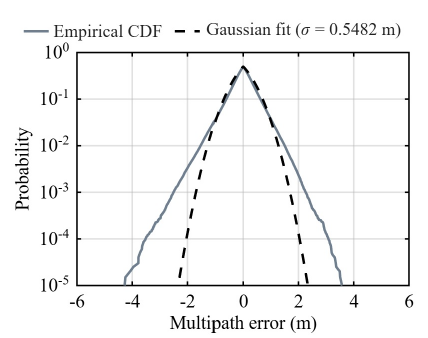}
    \caption{Global folded cumulative distribution functions (FCDF) for multipath errors. The solid grey curve shows the empirical FCDF, $\min\{F(s),1-F(s)\}$, where $F$ is the empirical CDF, and the black dashed line is the fitted one.}
    \label{fig:GNSS_LOGCDF_GLOBAL}
\end{figure}

\textbf{Fig}. \ref{fig:GNSS_GLOBAL} shows the global distribution of the multipath error, i.e., pooling all samples, in terms of the empirical probability density (left) and the quantile--quantile (QQ) plot (right). \textbf{Fig}. \ref{fig:GNSS_LOGCDF_GLOBAL} shows the global distribution in terms of folded cumulative distribution function (FCDF). The FCDF is obtained by taking the minimum between CDF and 1--CDF. The results suggest that the global distribution is not close to a Gaussian model. In the tails, the fitted Gaussian model underestimates the true probability of large errors. This motivates grouping samples according to elevation and C/N0 and evaluating the Gaussian fit within each group.

\begin{figure}[!t]
    \centering
    \includegraphics[width=0.9\columnwidth,trim=0 3pt 0 0,clip]
    {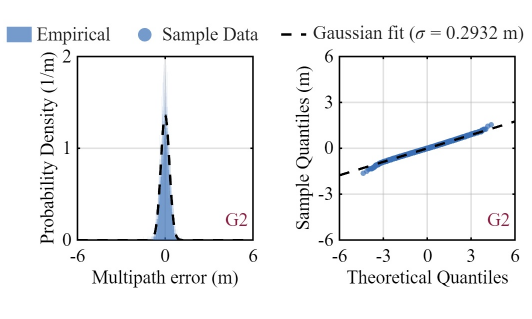}\par
    \includegraphics[width=0.9\columnwidth,trim=0 0 0 3pt,clip]
    {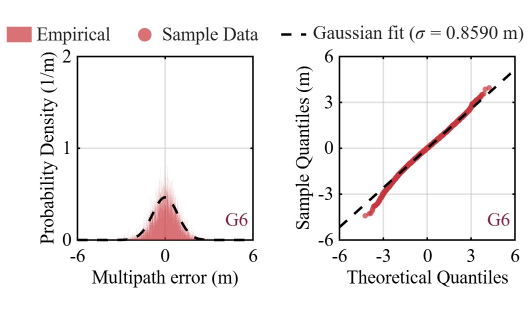}

    \caption{Distributions of GNSS multipath errors
    for G2 (top row) and G6 (bottom row). Each row shows the empirical
    and fitted Gaussian densities on the left and a QQ plot on the
    right.}
    \label{fig:GNSS_groups}
\end{figure}

\begin{figure}[!t]
    \centering
    \includegraphics[width=0.9\columnwidth]
    {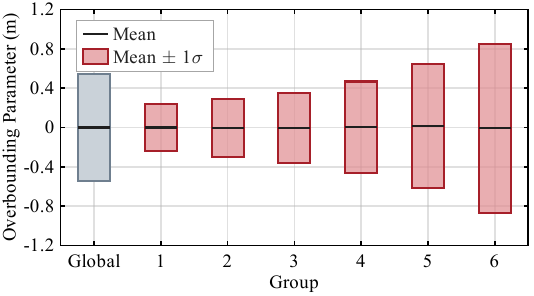}
    \caption{Calibrated means and standard deviations of GNSS multipath errors for the global model and for each group.}
    \label{fig:GNSS_test}
\end{figure}

\begin{figure}[!t]
\centering
\includegraphics[width=0.8\columnwidth]{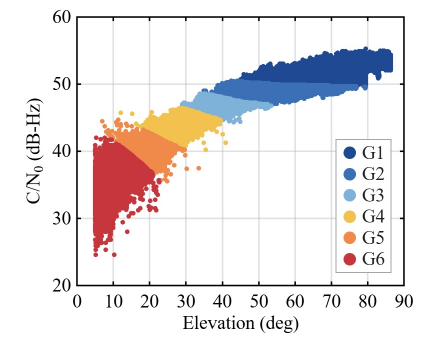}
\caption{Distribution of each group in the elevation-C/N0 plane.}
\label{fig:GNSS_Yspace}
\end{figure}

\textbf{Fig}. \ref{fig:GNSS_groups} shows G2 and G6 as representative small- and large-error groups, respectively. For G2 (i.e., QI=2), both the density and the QQ plot demonstrate close agreement with the fitted Gaussian. For G6, the fitted variance is larger than that of the global model, and the Gaussian fit is closer to the empirical distribution than the global fit. \textbf{Fig}. \ref{fig:GNSS_test} then compares the fitted means and standard deviations for each QI. The result shows two advantages of the conditional error model over the global one. First, it provides tighter and more accurate error models for most samples. For G1--G4, the fitted standard deviations are smaller than the global value; therefore, the global model overestimates their errors and is overly conservative. Second, it avoids underestimating the errors of the most challenging samples in G5 and G6; the global model would understate the probability of large errors for these samples, which may pose a risk in safety-critical applications. Finally, \textbf{Fig}. \ref{fig:GNSS_Yspace} visualizes the QI predictor by showing the samples in the elevation-C/N0 plane, colored by their QIs. This result will be further analyzed in Section \ref{sec:III-D}.

\subsection{Errors of Learning-based Relative Pose Regression}
\label{secIII-B}

This example considers a learning-based estimation system, i.e., Reloc3r~\cite{Dong_2025_CVPR}, a transformer-based network that estimates the relative camera pose between two images. We model the distribution of its relative pose estimation errors, particularly the three misalignment angles. Unlike the GNSS example, the QI predictor does not take hand-crafted physical features as inputs. Instead, it uses internal variables of Reloc3r, in which each image is encoded into a 1024-dimensional feature vector. This example therefore extends the evaluation in two aspects: the input space of the QI predictor is high-dimensional, and the error is modeled both as a scalar and as a vector. 

We use the DSLR images from ScanNet++ \cite{Yeshwanth_2023_ICCV}. It includes multiple scenes. For each scene, we construct multiple image pairs: each image frame is paired with up to 15 preceding and 15 following frames in the camera sequence. For each pair, Reloc3r with its official pretrained weights predicts the relative rotation matrix $\mathbf{R}$. By using the ground truth $\mathbf{R}_{\mathrm{gt}}$, the three misalignment angles are computed and expressed as a vector $\boldsymbol{\psi} = [\psi_x,\psi_y,\psi_z ]^\top$. With Reloc3r frozen, the features before its prediction head are pooled into a 1024-dimensional vector for each image, and the vectors of the two images are concatenated into a 2048-dimensional latent representation of the image pair, which serves as the input to the QI predictor. The dataset includes about 15 million image pairs from 1,006 scenes. A random subset of 3.11M pairs forms the training set, and the remaining pairs form the calibration set. We first model the scalar component $\psi_x$ and then extend the analysis to the vector $\boldsymbol{\psi}$. A separate QI predictor is trained for each setting, with $K=12$ in both cases. And for each setting, the training phase takes  $\sim$2.0 hours on a server with a GeForce RTX 5090 GPU (32GB).

\medskip
1) \textit{Scalar Error}

\begin{figure}[!t]
    \centering
    \includegraphics[width=0.9\columnwidth]
    {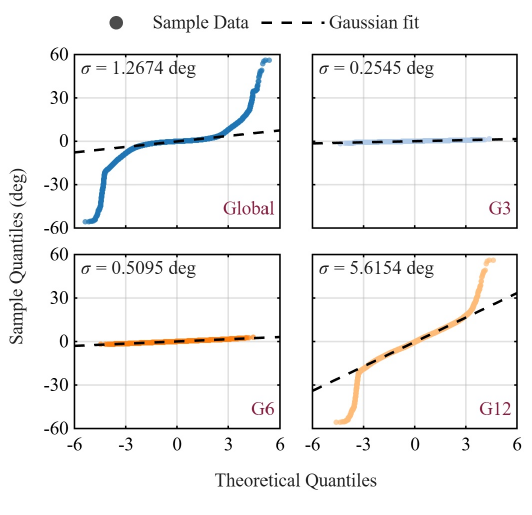}
    \caption{QQ plots of the misalignment angle error $\phi_x$ for the global distribution (Global) and for three groups G3, G6, and G12.}
    \label{fig:RPR_1_groups}
\end{figure}

\begin{figure}[!t]
    \centering
    \includegraphics[width=0.9\columnwidth]
    {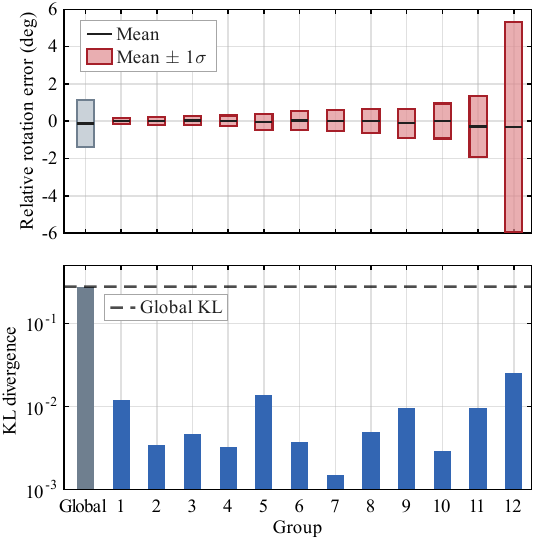}
    \caption{Statistics for the misalignment angle error $\phi_x$. Upper: fitted mean $\mu$ (horizontal line) and range $[\mu-\sigma,\mu+\sigma]$ (box) for Global and G1--G12. Lower: KL divergences between the empirical distributions and the fitted Gaussians; the dashed line marks the Global value.}
    \label{fig:RPR1_test}
\end{figure}

Fig.~\ref{fig:RPR_1_groups} compares the global distribution of the angle error $\psi_x$ with the distributions for three representative groups. Note that the QIs are labeled G1 through G12 in asscending order of the fitted variances. The global distribution deviates obviously from the fitted Gaussian in both tails. In contrast, the small- and medium-error group G3 and G6 agree closely with their fitted Gaussian. G12 contains many of the largest errors, and its fitted variance is larger than that of the global model. \textbf{Fig}. \ref{fig:RPR1_test} compares the fitted means and standard deviations of each group and also compares the KF divergence \eqref{eq:gaussianity_loss}. As shown in this figure, the fitted standard deviations vary from 0.16 to 5.62 degrees across the groups. Meanwhile, the KL divergence for each group is all much lower than the global one. These results show that the QIs predicted from image-pair features separate samples with substantially different error behaviors, and that the errors within each group are much better represented by a Gaussian than the global distribution.

\medskip
2) \textit{Multivariate Error}

\begin{figure*}[t]
    \centering
    \includegraphics[width=0.9\textwidth]
    {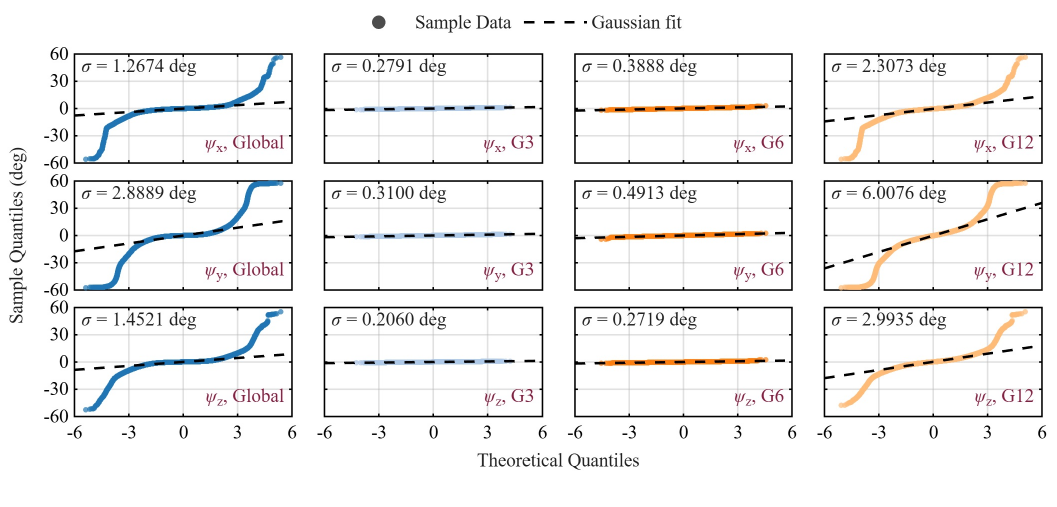}
    \caption{Marginal QQ plots of the relative rotation errors. The columns correspond to the pooled errors (Global) and groups G3, G6, and G12, and the rows correspond to the error components $\psi_x$, $\psi_y$, and $\psi_z$.}
    \label{fig:RPR_3_groups}
\end{figure*}

\begin{figure*}[t]
    \centering
    \includegraphics[width=0.8\textwidth]
    {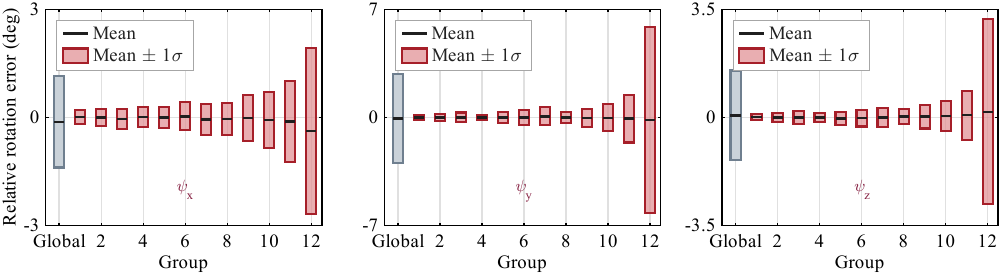}
    \caption{Fitted means and $\pm1\sigma$ ranges of the relative rotation error components $\psi_x$, $\psi_y$, and $\psi_z$ for G1--G12 and the pooled errors (Global).}
    \label{fig:RPR_test}
\end{figure*}

We then use the same image-pair features to model the three misalignment angles together. At each training iteration, the error vector $\boldsymbol{\psi}$ are projected onto 32 newly sampled random directions, and the distribution-matching loss encourages the projected errors within each group to approach Gaussian; this will promote a multivariate Gaussian distribution for $\boldsymbol{\psi}$ after a large number of iterations. \textbf{Fig}. \ref{fig:RPR_3_groups} shows the marginal QQ plots of $\psi_x$, $\psi_y$, and $\psi_z$ for the global distribution and the distributions of three groups. All three components of G3 and G6 agree closely with their fitted Gaussians, whereas G12 has a larger error spread. Note that the QIs are labeled in ascending order of the fitted variances for $\psi_x$. \textbf{Fig}. \ref{fig:RPR_test} shows the fitted means and standard deviations for each group and for the global distribution. These results suggest that a single QI can capture the error scales of all three components simultaneously, which enables a joint multivariate error model for each operating condition.

\subsection{Prediction Errors of a World Model}
\label{secIII-C}

The third example considers a prediction system, i.e., the LeWM world model~\cite{maes2026leworldmodel}, applied to the Push-T task, in which an agent pushes a T-shaped object on a 2-D plane.  Using past images and pushing actions, LeWM predicts the future latent embeddings of the scene, from which the pose of the T-shaped object is obtained using a decoder that we train on the frozen LeWM embeddings. We consider the errors of the predicted orientation $\alpha$ and position $[x,y]$ of the object. The QI predictor takes as inputs the predicted embeddings of LeWM, which are internal variables of the system. Compared with the previous examples, this example considers time-correlated errors. For the two settings below, a QI predictor with $K=12$ is trained separately. The training phase for each takes $\sim$1.5 hours.

\medskip
1) \textit{Multi-Step Orientation Error}

\begin{figure}[t]
    \centering
    \includegraphics[width=0.95\columnwidth]
    {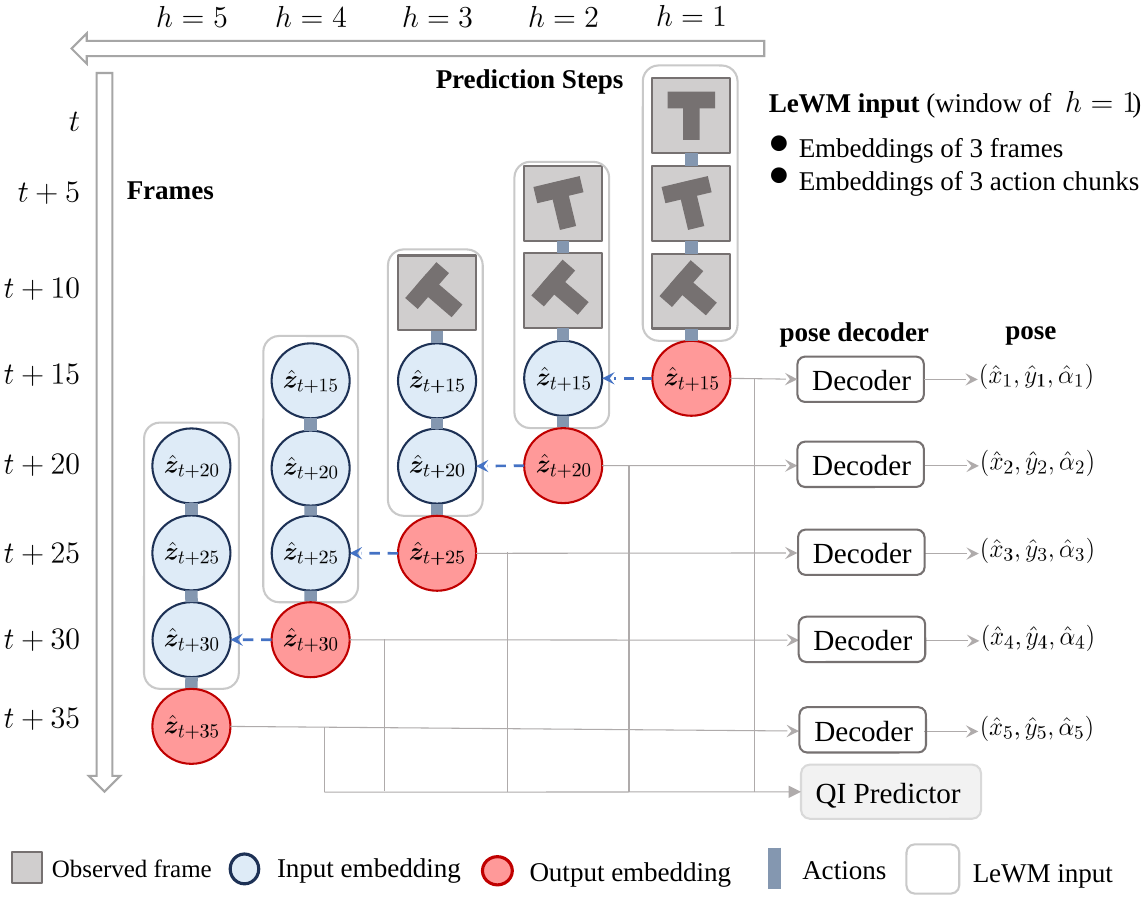}
    \caption{Five-step prediction process of the LeWM world model and the inputs to the QI predictor}
    \label{fig:pust_T}
\end{figure}

\textbf{Fig}.~\ref{fig:pust_T} illustrates the five-step prediction process of LeWM and presents the inputs to the QI predictor. At the first step $h=1$, LeWM takes as inputs the observed images at frames $t$, $t+5$, and $t+10$, together with 15 consecutive actions, and predicts the embeddings of the scene at frame $t+15$. These embeddings are then decoded into the object pose $(\hat{x}_1,\hat{y}_1,\hat{\alpha}_1)$ at frame $t+15$. At the second step $h=2$, the input window is shifted forward by 5 frames: LeWM takes the observed images at $t+5$ and $t+10$ and the embeddings predicted at the first step for frame $t+15$, together with the corresponding actions, and predicts the embeddings at frame $t+20$, which are decoded into $(\hat{x}_2,\hat{y}_2,\hat{\alpha}_2)$. The subsequent steps proceed in the same manner, with each new prediction fed back into the model, so that the five-step rollout yields predictions at frames $t+15$, $t+20$, $\ldots$, $t+35$. At each step $h$, the predicted orientation error $\Delta\alpha_h$ is computed using the ground truth. The five errors form the orientation error vector $\Delta\boldsymbol{\alpha}$. The QI predictor takes as inputs the five predicted embeddings.

\begin{figure}[t]
    \centering
    \includegraphics[width=0.8\columnwidth]
    {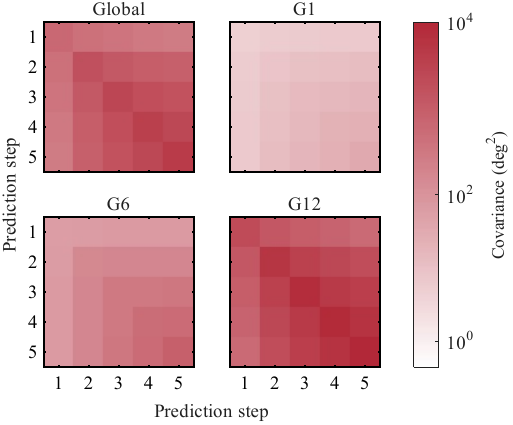}
    \caption{Fitted covariance matrices of the orientation prediction errors across five steps. The panels show Global, G1, G6, and G12 in reading order.}
    \label{fig:wm_theta5_correlation}
\end{figure}

We construct a dataset with 8 million 5-step windows from 95,000 trajectories. Within each trajectory, we randomly select $\sim$80\% of the windows for training and use the remaining for calibration. To capture the temporal correlation of the errors, a QI predictor is trained to model the 5-step orientation 
errors jointly, so that each error vector $\Delta\boldsymbol{\alpha}$ is associated with a single QI. The QIs are labeled G1 through G12 in ascending order of the fitted standard deviation of the first-step angular error $\Delta\alpha_1$. \textbf{Fig}.~\ref{fig:wm_theta5_correlation} shows the fitted covariance matrices of the 5-step
orientation errors for the global model and three groups. The results show that the groups exhibit clearly different error behaviors. In addition, they show positive correlations between the errors at different steps and that the error variance increases toward later steps. This reflects the accumulation of errors along the time steps. 

\begin{figure*}[t]
    \centering
    \includegraphics[width=0.9\textwidth]
    {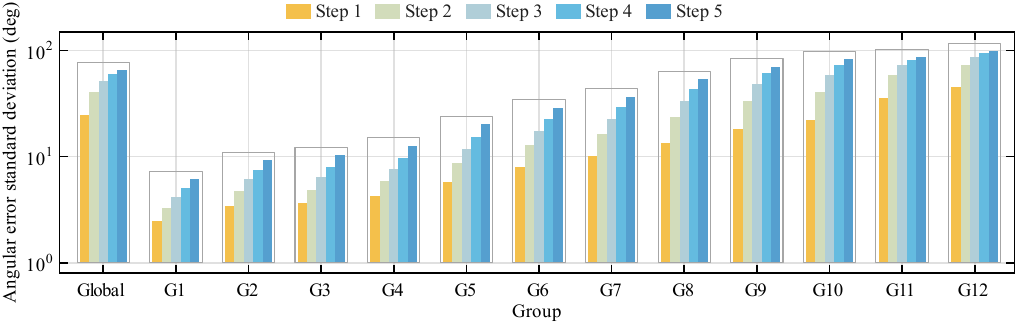}
    \caption{Fitted standard deviations of the orientation prediction errors for the pooled errors (Global) and G1--G12 at each step.}
    \label{fig:wm_theta5_group_statistics}
\end{figure*}

\begin{figure*}[t]
    \centering
    \includegraphics[width=0.8\textwidth]
    {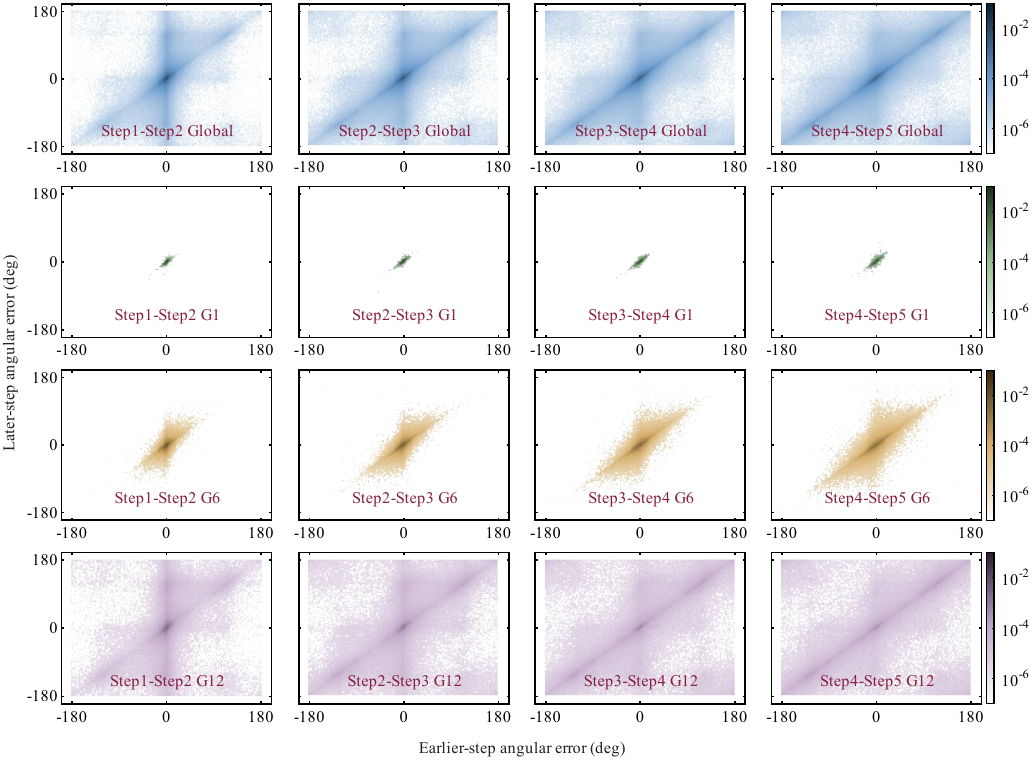}
    \caption{Scatter plots of the orientation prediction errors at adjacent steps. The rows correspond to Global, G1, G6, and G12, and the columns correspond to the step pairs $(1,2)$, $(2,3)$, $(3,4)$, and $(4,5)$.}
    \label{fig:wm_theta5_adjacent_scatter}
\end{figure*}

\textbf{Fig}.~\ref{fig:wm_theta5_group_statistics} compares the fitted standard deviations for different steps and groups. The standard deviations differ substantially across groups at every step, e.g., from $6.12^\circ$ in G1 to $99.48^\circ$ in G12 at step 5. \textbf{Fig}.~\ref{fig:wm_theta5_adjacent_scatter} further presents the joint distributions of the errors at adjacent steps. It shows that G1 and G6 are closer to bivariate Gaussian distributions than the global errors. G12, which collects the samples with the largest errors, still deviates from a Gaussian distribution. These results suggest the effectiveness of our \textsc{condUQ} approach.

\medskip
2) \textit{One-Step Position Error}

We also model the two-dimensional position errors of one-step predictions. Given the images at frames $i$, $i+5$, and $i+10$ and the corresponding actions, LeWM predicts the object pose at frame $i+15$. The differences between the predicted and ground-truth positions along the two image axes give the error vector $\boldsymbol{e}=[\Delta x_1,\Delta y_1]^{\top}$. A separate QI predictor takes the 192-dimensional predicted embedding of this step as input and models the two error components jointly.

\begin{figure*}[!t]
    \centering
    \includegraphics[width=0.9\textwidth]
    {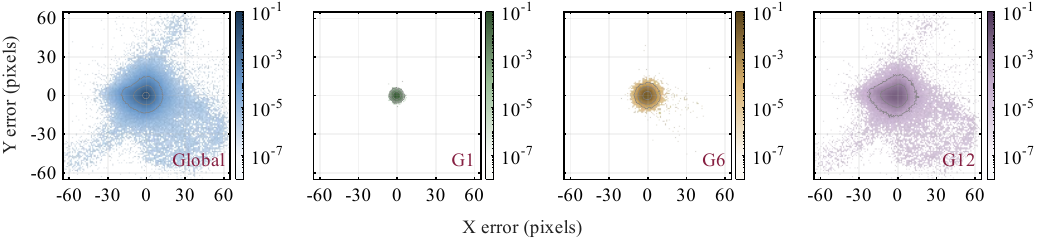}
    \caption{Scatter plots of the one-step position errors $(e_x,e_y)$ for the pooled errors (Global) and groups G1, G6, and G12.}
    \label{fig:WM_XY}
\end{figure*}

\textbf{Fig}.~\ref{fig:WM_XY} shows the empirical joint distributions of the one-step position errors for the global model and for three groups, G1, G6, and G12. The global errors spread over a wide range and exhibit heavy tails. In contrast, the errors in G1 and G6 approximately follow bivariate Gaussian distributions and are much more concentrated than the global errors. G12 absorbs most of the large errors, so it has a much larger spread and heavy tails similar to those of the global errors. By isolating these samples in a dedicated group, \textsc{CondUQ} prevents them from inflating the error models of the other groups. These results demonstrate the benefit of conditional error modeling and the effectiveness of \textsc{CondUQ} for multivariate errors.

\subsection{Consistency of the Learned QIs with Physical Principles}
\label{sec:III-D}

For the \textsc{condUQ} approach, the QI predictor is trained only with the distribution-matching objective, without any physical knowledge of the error sources. It is therefore natural to ask whether the learned QIs reflect the physical factors that govern the error behavior. We examine this question using the GNSS example and the scalar relative rotation error example.

\medskip
1) \textit{GNSS multipath errors}

Both inputs of the QI predictor in this example have direct physical interpretations. The satellite elevation describes the observation geometry: the signals from low-elevation satellites travel closer to the ground and are more likely to be reflected by surrounding objects. The carrier-to-noise density ratio C/N0 indicates the received signal quality, which degrades when reflected signals interfere with the direct signal. Both factors are therefore expected to be associated with larger multipath errors when their values are low. As presented in Fig.~\ref{fig:GNSS_Yspace} in Sec. \ref{secIII-A}, the large-error groups occupy mainly the region where both elevation and C/N0 are low, whereas the small-error groups concentrate where both are high. This pattern is consistent with the physical understanding of multipath effects, and it is learned from data without predefined elevation or C/N0 thresholds.

\medskip
2) \textit{Relative rotation errors of Reloc3r} 

In this example, the QI predictor takes as input a 2048-dimensional feature vector extracted by Reloc3r from each image pair, whose individual components are challenging to interpret directly. From multi-view geometry, however, it is well understood that the accuracy of relative pose estimation depends on the number of reliable feature correspondences between the two images and on their spatial distribution: few matches or matches concentrated in a small image region lead to a poorly conditioned estimation problem and hence a large uncertainty. Hence, we construct a physically meaningful reference that accounts for both factors. For each image pair, we use SuperPoint~\cite{detone2018superpoint} to detect keypoints and extract their descriptors, and utilize LightGlue~\cite{lindenberger2023lightglue} to match the keypoints across the two images. The matches are then used to estimate the relative pose and its covariance matrix, whose first diagonal entry $\mathrm{cov}_{11}$ is the estimated variance of the rotation component about the $x$-axis. Its square root, $\sigma_{\mathrm{geo}}=\sqrt{\mathrm{cov}_{11}}$, thus provides a per-pair measure of the geometric uncertainty of the rotation component corresponding to the misalignment angle $\psi_x$. Note that $\sigma_{\mathrm{geo}}$ is computed independently of Reloc3r and is used only for this analysis, not for training the QI predictor.

\begin{figure}[!t]
    \centering
    \includegraphics[width=0.8\linewidth]
    {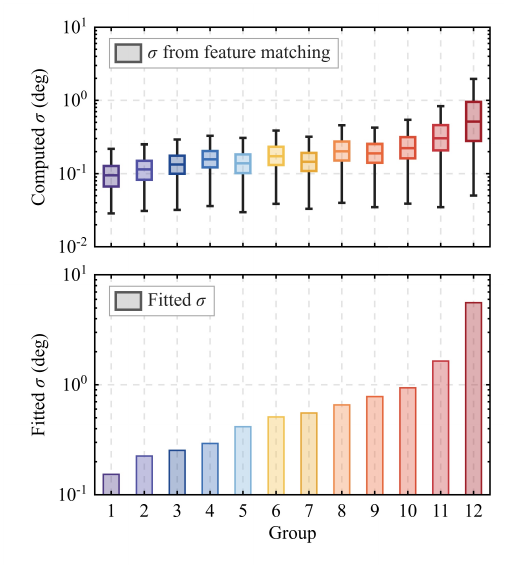}
    \caption{Median geometric uncertainty $\sigma_{\mathrm{geo}}$ obtained from feature matching and fitted standard deviation of the rotation error $\phi_x$ for each group.}
    \label{RPR_1_feature}
\end{figure}

Fig.~\ref{RPR_1_feature} compares the median of $\sigma_{\mathrm{geo}}$ within each group with the fitted standard deviation of $\psi_x$. Groups with larger fitted standard deviations generally exhibit larger median of $\sigma_{\mathrm{geo}}$. That is, the image pairs that the QI predictor assigns to large-error groups are also those with fewer or less well-distributed feature matches. This agreement suggests that, although the QIs are learned from high-dimensional and abstract network features, they capture the geometric factors that physically determine the accuracy of relative pose estimation.

\subsection{Applications of Uncertainty Quantification Results}

Accurate uncertainty quantification can benefit a system in several ways. First, it determines when an output can be safely used and how much confidence can be placed in it, which is essential for safety-critical decisions. Second, it helps diagnose the performance of a measurement, estimation, or prediction system by revealing the operating conditions under which large errors are likely to appear. Such diagnostics provide guidance for improving the system, for example, by indicating which conditions require better sensors, more training data, or refined models. In addition, the quantified uncertainty can serve as weights when the outputs are combined in downstream tasks such as information fusion and state estimation. In this section, we illustrate the last application using GPS positioning, where the QI-conditional error variances learned by \textsc{CondUQ} are used to weight pseudorange measurements. Other applications, such as uncertainty-aware planning with world models and decision-making in autonomous driving \cite{khaitan2021safe}, will be investigated in our future work.

\textbf{Experimental setup.} We consider pseudorange-based GPS differential positioning as an example. We compare a weighted least-squares (WLS) estimator, in which the measurements are weighted by the learned error variances, with an ordinary least-squares (OLS) estimator, which uses the global error model. Two IGS receivers, ACSX and AR27, are used, with a distance of 13 km. ACSX serves as the reference station with known coordinates, and the position of AR27 is estimated. 24-hour data on January 1, 2025, with a sampling interval of 1 s, are used to evaluate the positioning performance.

\begin{figure}[!t]
    \centering
    \includegraphics[width=\columnwidth]
    {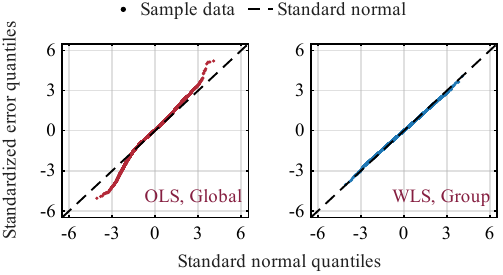}
    \caption{QQ plots of the standardized positioning errors along the ECEF $X$ axis. Left: OLS with the global error model. Right: WLS with the QI-conditional error models of \textsc{CondUQ}.}
    \label{fig:position_qq}
\end{figure}

\textbf{Positioning algorithm.} At each epoch and for each satellite observed by both receivers, we form a between-receiver single difference by subtracting the pseudorange measured by ACSX from that measured by AR27. This differencing removes most of the error sources, leaving multipath errors as the major one. The multipath errors of two receivers are independent of each other. The single differences are used to estimate the position of AR27 using WLS and OLS. Using the same set of observations, OLS and WLS estimate the three-dimensional position of AR27 and the between-receiver clock offset. OLS assigns equal weights to all single differences. In contrast, for WLS, each undifferenced observation is assigned the variance of the calibrated error model associated with its QI. Assuming that the errors at the two stations are independent, the variance of each single difference is the sum of the two station-specific variances, and its weight is the inverse of this sum. For OLS and WLS, the predicted position covariances are respectively
\begin{equation}
\mathbf{P}_{\mathrm{OLS}}=\sigma_{\mathrm{g}}^2(\mathbf{H}^{\top}\mathbf{H})^{-1}, \quad \mathbf{P}_{\mathrm{WLS}} = (\mathbf{H}^{\top}\mathbf{W}\mathbf{H})^{-1}
\label{eq:WLS_OLS_COV}
\end{equation}
where $\mathbf{H}$ is the observation matrix, $\sigma_g$ is the variance of the single differences obtained using the global error model, and $\mathbf{W}$ is the diagonal weight matrix formed by the inverse single-difference variances that is obtained using \textsc{condUQ}. Table~\ref{tab:position_rms} compares the position errors from OLS and WLS. The result shows that WLS reduces the root-mean-square (RMS) of the positioning errors by 21.9\%, 22.3\%, and 17.9\% along the $X$, $Y$, and $Z$ axes, respectively, compared with OLS. Then, Fig.~\ref{fig:position_qq} shows the QQ plots of the standardized $X$-axis errors, where each position error is divided by the covariance predicted by the corresponding estimator, given in \eqref{eq:WLS_OLS_COV}. As shown in this figure, the standardized errors of WLS follow the standard normal reference line more closely than those of OLS, with smaller deviations in the tails. These results show that weighting the measurements by the QI-conditional variances not only improves the positioning accuracy along all three axes but also yields position errors whose distribution is better described by the predicted uncertainty.

\begin{table}[!t]
    \centering
    \caption{Root-Mean-Square (RMS) of the Positioning Errors Along the Three ECEF Axes}
    \label{tab:position_rms}
    \renewcommand{\arraystretch}{1.3}
    \begin{tabular*}{\columnwidth}{@{\extracolsep{\fill}}lccc@{}}
        \hline
        Method & $X\!\downarrow$ (m) & $Y\!\downarrow$ (m) & $Z\!\downarrow$ (m) \\
        \hline
        OLS (global error model) & 0.3459 & 0.4437 & 0.3990 \\
        WLS (\textsc{CondUQ}) & \textbf{0.2701} & \textbf{0.3448} & \textbf{0.3275} \\
        \hline
    \end{tabular*}
\end{table}

\subsection{Ablation Study on the Balance Regularization}

We use the five-step orientation prediction task of the world model example to examine the impact of the balance regularization term in~\eqref{eq:qi_training_loss} on the QI assignments. As mentioned in Sec.~\ref{secIII-C}, the five-step orientation errors of each prediction window are modeled jointly, and the QI predictor takes a 960-dimensional feature vector as the input. We fix $K=12$ and compare the QI assignments on the test set after 2,000 training iterations under two settings: (a) without the balance regularization ($\eta=0$) and (b) with the balance regularization whose weight $\eta$ decays linearly from $\eta_0=6$.

\begin{figure}[!t]
    \centering
    \includegraphics[width=0.8\columnwidth]{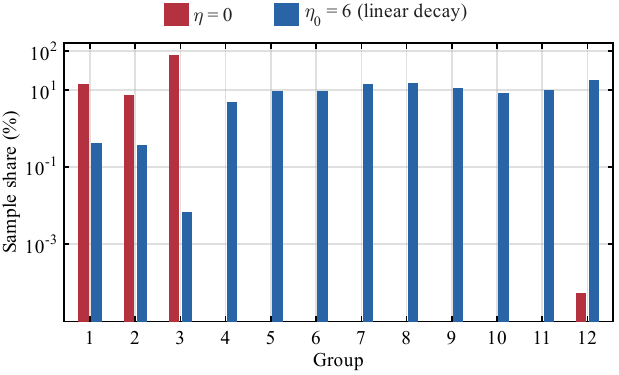}
    \caption{Proportions of samples assigned to each group with and without the balance regularization in the five-step angular prediction task. The groups obtained with $\eta_0=6$ are sorted in ascending order of the fitted variance of the first-step angular error, and the groups obtained with $\eta=0$ are shown in the order of their original indices. The proportions are plotted on a logarithmic scale, and only nonempty groups are shown.}
    \label{fig:wm_balancing_ablation}
\end{figure}

Fig.~\ref{fig:wm_balancing_ablation} compares the number of samples within each group under both settings. As shown in Fig.~\ref{fig:wm_balancing_ablation}, without the balance regularization, only 4 of the 12 groups receive samples. The largest group contains 78.43\% of the samples, and one of the nonempty groups contains only a single sample. In this case, the QI predictor collapses to a few groups, and most samples with heterogeneous error behavior are pooled together, which undermines the purpose of conditional error modeling. With the balance regularization, all 12 groups receive samples from the training set, and the largest group takes only 17.51\% of the samples. Therefore, the balance regularization effectively prevents group collapse and makes full use of the groups. In the Appendix, we further show the sensitivity analysis results over the number of iterations, the number of groups, and the decay mechanism of the balance regularization weight.

\section{Conclusion}

This paper proposed \textsc{CondUQ}, a calibrated conditional uncertainty quantification framework that can be built on top of any model- or learning-based measurement, estimation, or prediction system. Instead of describing the output error with a single global distribution, \textsc{CondUQ} learns a discrete quality index (QI) from the observable inputs and internal variables of the system, by using a distribution-matching objective that requires no quality labels. This objective encourages the errors within each QI to follow a common, approximately Gaussian distribution. A separate error model is then fitted and calibrated for each QI on held-out data; during online operation, the error distribution of a new output can be obtained from observable variables alone, without access to the ground truth.

\textsc{CondUQ} was evaluated on three systems of different types: GNSS multipath errors conditioned on two physical features, attitude errors of a transformer-based relative pose estimation model conditioned on high-dimensional network features, and prediction errors of a world model conditioned on the latent embeddings. In all three examples, the QI-conditional error distributions agreed much better with the Gaussian model than the global distribution, for both scalar and multivariate errors. The learned QIs were also found to be consistent with physical principles: large-error groups corresponded to low elevation and low C/N0 in the GNSS example, and to large geometric uncertainty derived from feature matching in the relative pose regression example. All the results suggest the effectiveness of \textsc{condUQ} on achieving accurate and reliable uncertainty quantification and also show its efficiency.

Several directions remain for future work. First, \textsc{CondUQ} characterizes the errors under the operating conditions covered by the calibration data, and its extension to out-of-distribution inputs will be investigated. In addition, we will explore the use of the resulting uncertainty in other downstream applications, such as uncertainty-aware task planning with world models and decision-making in autonomous driving.

\appendices
\section*{Appendix}

\subsection{Sensitivity Analysis}
\label{app:balancing_sweep}

This appendix provides results of a sensitivity analysis with respect to the number of training iterations, the number of groups, and the setting of the balance regularization weight $\eta$ in~\eqref{eq:qi_training_loss}. We use the scalar relative rotation error $\phi_x$ of the relative pose regression system for the comparison. The numbers of iterations are 500, 1,000, and 2,000, and the numbers of groups are $K=6$ and $12$. Three settings of the balance regularization are compared: no regularization ($\eta=0$), a constant weight ($\eta=3$), and a weight that decays linearly from $\eta_0=3$.

Two metrics are computed on the calibration set. The first is the normalized entropy $\mathcal{H}_{\mathrm{QI}}$ in~\eqref{eq:balance_entropy}, evaluated with the hard QI assignments of the samples, i.e., with $\bar{\pi}_k=n_k/N_{\mathrm{te}}$, where $n_k$ is the number of samples assigned to QI $=k$ and $N_{\mathrm{te}}$ is the total number of samples. Larger values of $\mathcal{H}_{\mathrm{QI}}$ indicate more balanced assignments. The second metric measures how well the error distribution of each QI is represented by a Gaussian. Following~\eqref{eq:projected_kl}, the KL divergence for QI $=k$ is
\begin{equation}
D_k=\sum_{b=1}^{B}\widehat{p}_{kb}\log\frac{\widehat{p}_{kb}}{\widehat{g}_{kb}},
\end{equation}
where $\widehat{p}_{kb}$ and $\widehat{g}_{kb}$ are the probability masses of the empirical distribution and the fitted Gaussian in the $b$-th bin, respectively, and $B=10{,}000$ bins are used. Let $\mathcal{K}_{\mathrm{valid}}$ denote the set of QIs with valid KL values. The average KL divergence is
\begin{equation}
\overline{D} = \frac{1}{|\mathcal{K}_{\mathrm{valid}}|}\sum_{k\in\mathcal{K}_{\mathrm{valid}}}D_k.
\end{equation}
Unlike the training loss $\mathcal{L}_{\mathrm{G}}$ in~\eqref{eq:gaussianity_loss}, which is computed with soft assignments, $\overline{D}$ is computed with the hard assignments of the samples and directly measures how close the QI-wise error distributions are to Gaussian, with smaller values indicating a closer match.

\begin{table}[t]
\centering
\caption{Sensitivity of the QI Assignments and Error Models to the Balance Regularization for the Relative Rotation Error $\phi_x$}
\label{tab:rpr1_balancing_ablation}
\renewcommand{\arraystretch}{1.15}
\begin{tabular*}{\columnwidth}{@{\extracolsep{\fill}}cccccc@{}}
\hline
Iterations & $K$ & $\eta_0$ & Decay & $\mathcal{H}_{\mathrm{QI}}\uparrow$ & $\overline{D}\downarrow$ \\
\hline
500 & 6 & 0 & --- & 0.4539 & 0.0342 \\
 &  & 3 & --- & \textbf{0.9996} & \textbf{0.0186} \\
 &  & 3 & Linear & 0.8273 & 0.0230 \\
\hline
500 & 12 & 0 & --- & 0.3343 & 0.0866 \\
 &  & 3 & --- & \textbf{0.9921} & \textbf{0.0109} \\
 &  & 3 & Linear & 0.7171 & 0.0133 \\
\hline
1,000 & 6 & 0 & --- & 0.5064 & 0.0285 \\
 &  & 3 & --- & \textbf{0.9983} & 0.0162 \\
 &  & 3 & Linear & 0.8508 & \textbf{0.0071} \\
\hline
1,000 & 12 & 0 & --- & 0.3607 & 0.0268 \\
 &  & 3 & --- & \textbf{0.9809} & \textbf{0.0049} \\
 &  & 3 & Linear & 0.7299 & 0.0083 \\
\hline
2,000 & 6 & 0 & --- & 0.5647 & 0.0242 \\
 &  & 3 & --- & \textbf{0.9992} & 0.0149 \\
 &  & 3 & Linear & 0.7630 & \textbf{0.0053} \\
\hline
2,000 & 12 & 0 & --- & 0.4251 & 0.0193 \\
 &  & 3 & --- & \textbf{0.9855} & \textbf{0.0037} \\
 &  & 3 & Linear & 0.6665 & 0.0048 \\
\hline
\end{tabular*}
\end{table}

Table~\ref{tab:rpr1_balancing_ablation} summarizes the results, from which we can make three observations. First, the balance regularization effectively improves the balance of the assignments. Without this term, $\mathcal{H}_{\mathrm{QI}}$ ranges from 0.33 to 0.56, indicating that most samples are concentrated in a few QIs. With a constant weight $\eta=3$, $\mathcal{H}_{\mathrm{QI}}$ exceeds 0.98 in all settings, and with a linearly decaying weight, it lies between 0.67 and 0.85.

Second, the balance regularization improves the Gaussianity of the QI-conditioned error distributions. In every setting, both the constant and the linearly decaying weights yield a lower $\overline{D}$ than training without regularization. Compared with no regularization, the better of the two weighting schemes reduces $\overline{D}$ by 45.6\%--87.4\%, with the largest reduction for $K=12$ after 500 iterations, where $\overline{D}$ decreases from 0.0866 to 0.0109. Thus, by preventing group collapse, the balance regularization does not trade the quality of the error models for balanced assignments but improves both.

Third, the constant and linearly decaying weights perform comparably. The constant weight yields the most balanced assignments in all settings and the lowest $\overline{D}$ for $K=12$. The linearly decaying weight yields the lowest $\overline{D}$ for $K=6$ after 1,000 and 2,000 iterations, which is consistent with its design: a strong regularization in the early stage prevents group collapse, and a weaker regularization in the later stage allows the distribution-matching objective to dominate and refine the groups, at the cost of less balanced assignments. In addition, $\overline{D}$ decreases as the number of training iterations increases in all settings.

\end{document}